\documentclass[11pt,letterpaper]{article}

\usepackage{emnlp2017}

\usepackage{times}
\usepackage{relsize}
\usepackage[T1]{fontenc} 
\usepackage[latin1]{inputenc}
\usepackage[english]{babel}

\usepackage{epsfig}
\usepackage{graphicx}
\usepackage{wrapfig}
\usepackage[belowskip=-2pt,aboveskip=3pt,font=small]{caption}
\usepackage[belowskip=0pt,aboveskip=3pt,font=small]{subcaption}
\usepackage{amsmath, amsthm, amssymb}
\usepackage{textcomp}
\usepackage{stmaryrd}
\usepackage{upgreek}
\usepackage{bm}
\usepackage{cases}
\usepackage{mathtools}
\usepackage{arydshln}
\usepackage{multirow}

\usepackage{algorithm}
\usepackage{algpseudocode}

\usepackage{multirow}
\usepackage{rotating}
\usepackage{booktabs}

\usepackage{enumitem}
\usepackage[olditem,oldenum]{paralist}

\usepackage{alltt}
\usepackage{listings}

\usepackage{mysymbols}

\usepackage{url}
\usepackage{xspace}
\usepackage{comment}
\usepackage{color}
\usepackage{xcolor}
\usepackage{afterpage}
\usepackage{framed}
\usepackage{fancybox}
\usepackage{cuted}
\usepackage{caption}
\usepackage{flushend}

\belowdisplayskip \abovedisplayskip

\newlength{\sectionReduceTop}
\newlength{\sectionReduceBot}
\newlength{\subsectionReduceTop}
\newlength{\subsectionReduceBot}
\newlength{\abstractReduceTop}
\newlength{\abstractReduceBot}

\newlength{\eqnReduceTop}
\newlength{\eqnReduceBot}

\newlength{\horSkip}
\newlength{\verSkip}

\newlength{\figureHeight}
\usepackage{latexsym}
\usepackage{amssymb}
\usepackage{pifont}
\newcommand{\cmark}{\ding{51}}%
\newcommand{\xmark}{\ding{55}}%
\usepackage[normalem]{ulem}

\usepackage{booktabs}

\usepackage{tabu} 
\usepackage{subcaption}

\newcommand{\Qbot}{\textsc{Q-bot}\xspace}
\newcommand{\Abot}{\textsc{A-bot}\xspace}

\newcommand{\tnt}{Task \& Talk\xspace}

\newcommand{\reffig}[1]{Fig.~\ref{#1}}
\newcommand{\refsec}[1]{Sec.~\ref{#1}}
\newcommand{\reftab}[1]{Tab.~\ref{#1}}

\newcommand{\ad}[1]{\textcolor{black}{#1}}
\newcommand{\sk}[1]{\textcolor{black}{#1}}

\newcommand{\sml}[1]{\textcolor{black}{#1}}
\newcommand{\FIXME}[1]{\textbf{\textcolor{red}{#1}}}

\emnlpfinalcopy

\def\emnlppaperid{505}

\title{Natural Language Does Not Emerge `Naturally' in Multi-Agent Dialog}

\author{
Satwik Kottur$^1$ \and
Jos\'e M.F. Moura$^1$ \and
Stefan Lee$^{2,3}$ \and
Dhruv Batra$^{3,4}$ \\
$^1$Carnegie Mellon University, $^2$Virginia Tech, $^3$Georgia Tech, $^4$Facebook AI Research\\
 }

\date{}

\begin{document}

\maketitle

\begin{abstract}
A number of recent works have proposed techniques for end-to-end learning of  communication protocols among cooperative multi-agent populations, and have simultaneously found the \emph{emergence of grounded human-interpretable language} in the protocols developed by the agents, learned without any human supervision!  
%

In this paper, using a \emph{\tnt} 
reference game between two agents as a testbed, 
we present a sequence of `negative' results
culminating in a `positive' one -- showing
that while most agent-invented languages are effective 
(\ie achieve near-perfect task rewards), they
are decidedly \emph{not} interpretable or compositional. 
In essence, we find that natural language does not emerge `naturally',
despite the semblance of ease of natural-language-emergence that one may gather from recent literature.
We discuss how it is possible to coax the invented languages to become more and more human-like and compositional by increasing restrictions on how two agents may communicate.
\end{abstract}

\vspace{\sectionReduceTop}
\section{Introduction}
\label{sec:intro}
\vspace{\sectionReduceBot}
One fundamental goal of artificial intelligence (AI) is the development of goal-driven dialog agents --
specifically, agents that can perceive their environment (through vision, audition, or other sensors),
and communicate with humans or other agents in natural language 
towards some goal.

While historically such agents have been based on \emph{slot filling}~\cite{lemon_eacl06},
the dominant paradigm today is neural dialog models~\cite{bordes_arxiv16,weston_arxiv16,serban_aaai16,serban_arxiv16}
trained on large quantities of data.
Perhaps somewhat counterintuitively, this current paradigm
treats dialog as a static supervised learning problem, rather than as the interactive agent learning problem that it naturally is.~Specifically,
a typical pipeline is to collect a large dataset of human-human dialog~\cite{lowe_sigdial15,visdial,guesswhat,mostafazadeh_arxiv17},
inject a machine in the middle of a dialog from the dataset, and supervise it to mimic the human response.
While this teaches the agent correlations between symbols, 
it does not convey the functional meaning of language,  
grounding (mapping words to physical concepts), compositionality (combining knowledge of simpler concepts to describe richer concepts), 
or aspects of planning (understanding the goal of the conversation).

An alternative paradigm that has a long history \cite{winograd1971procedures,kirby2014iterated} and is witnessing a
recent resurgence~\cite{wang2016learning,foerster_nips16,sukhbaatar2016learning,jorge_iclrw16,lazaridou_iclr17,havrylov_iclrw17,mordatch_arxiv17,visdial_rl} --
is situated language learning.
A number of recent works have proposed reinforcement learning techniques to learn communication protocols of agents situated in virtual environments
in a completely end-to-end manner -- from perceptual input (\eg pixels) to communication (discrete symbols without any pre-specified meanings) to action
(\eg signaling in reference games or navigating in an environment) --
and have simultaneously found the \emph{emergence of grounded
human-interpretable (often compositional) language} 
in the communication protocols developed by the agents, 
without any human supervision or pretraining, simply to succeed at the task. 


In this paper, we study the following question --
what are the conditions that lead to the emergence
of human-interpretable or compositional grounded language?
%
%
%
Our key finding is that
natural language \emph{does not emerge `naturally'}
in multi-agent dialog,
despite the semblance of ease of natural-language-emergence in multi-agent games that one may gather from recent literature.

Specifically, in a sequence of `negative' results culminating in a `positive' one,
we find that while agents always successfully invent communication protocols and languages to achieve their goals with
near-perfect accuracies, the invented languages are decidedly \emph{not} compositional, interpretable, or `natural'; and that it is possible to coax
the invented languages to become more and more human-like and compositional
by increasing restrictions on how two agents may communicate.

\textbf{Related work and novelty.}
The starting point for our investigation is the recent work of \citet{visdial_rl}, 
who proposed a cooperative reference game between two agents, where communication 
is necessary to accomplish the goal due to an information asymmetry. 
Our key contribution over \citet{visdial_rl} is an exhaustive study of the 
conditions that must be present before compositional grounded language emerges, 
and subtle but important differences in execution -- tabular Q-Learning (which does not scale) 
\vs REINFORCE (which does), and generalization to novel environments (not studied in prior work). 
We hope our findings shed more light into the interpretability of 
languages invented in cooperative multi-agent settings, 
place recent work in appropriate context, and 
inform fruitful directions for future work.
\vspace{\sectionReduceTop}
\section{The \emph{\tnt} Game}
\label{sec:setup}
\vspace{\sectionReduceBot}

\begin{figure}
	\centering
    \includegraphics[width=0.99\linewidth]{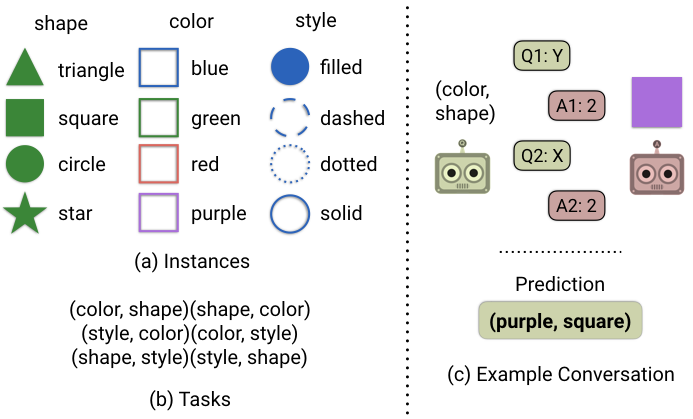}
    \caption{(a) \emph{\tnt}: The testbed for our study is cooperative 2-player game, \emph{\tnt}, grounded in 
  a synthetic world of objects with 4 \emph{shapes} $\times$ 4 \emph{colors} $\times$ 4 \emph{styles}.
  (b) \Qbot is assigned a \emph{task} -- to inquire about the state of an ordered pair of attributes.
  (c) An example gameplay between the two agents - \Qbot asks questions depending on the task which are answered by \Abot conditioned on the hidden instance visible to only itself. At the end, \Qbot makes a prediction of pair of attributes \emph{(purple, square)}.}
    \label{fig:setup_world}
\end{figure}

Our testbed is a cooperative reference game (\emph{\tnt}) between two agents, \Qbot
and \Abot. 
The game is grounded in a synthetic world of
objects comprised of three attributes 
-- \emph{color}, \emph{style}, and \emph{shape} -- each with four possible values for a total of $4\times4\times4=64$ objects.
\figref{fig:setup_world}a shows all the possible attribute values.

\emph{\tnt} plays out over multiple rounds of dialog.
At the start, \Abot is given an object unseen by \Qbot, \eg \emph{(green, dotted, square)}.
On the other side, \Qbot is assigned a task $G$ (unknown to \Abot) consisting of two attributes, \eg \emph{(color, style)}
and the goal is for \Qbot to discover these two attributes of the hidden object, through dialog with \Abot.
\sk{
Specifically, \Qbot and \Abot exchange utterances from finite vocabularies 
$V_Q$ and $V_A$ over two rounds, with \Qbot speaking first.}
The game culminates in \Qbot guessing a pair of attribute values, 
\eg \emph{(green, dotted)}, and both agents are rewarded identically 
based on the accuracy of \Qbot`s prediction.

Note that the \tnt game setting involves an informational asymmetry between the agents -- \Abot sees the object while \Qbot does not; similarly \Qbot knows the task while \Abot does not.
Thus, a two-way communication is necessary for success. Without this asymmetry, \Abot could simply convey the target attributes from the task without \Qbot having to speak. Such a setting has been widely studies in economics and game theory as the classic Lewis Signaling (LS) game \cite{lewis2008convention}. By necessitating \emph{dialog} between agents, we are able ground both $V_A$ and $V_Q$ in our final setting (\refsec{sec:nomem_min}).

\vspace{\sectionReduceTop}
\section{Modeling \Qbot and \Abot}
\label{sec:model}
\vspace{\sectionReduceBot}

\begin{figure*}[t]
  \centering
  \includegraphics[width=\textwidth]{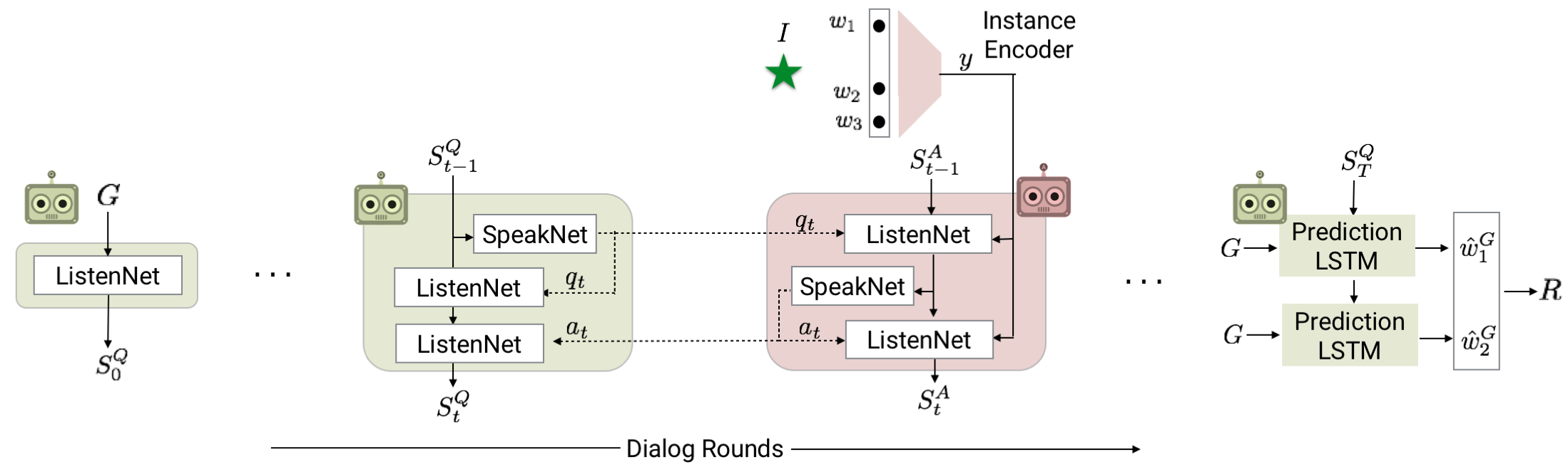}
  \caption{Policy networks for \Qbot and \Abot. At each round $t$ of dialog,
  (1) \Qbot generates a question $q_t$ from its speaker network conditioned on its state encoding $S^Q_{t-1}$,
  (2) \Abot encodes $q_t$ conditioned on instance $y$ encoded via instance encoder, updates its state encoding $S^A_t$, and generates an answer $a_t$,
  (3) \Qbot encodes $(q_t, a_t)$ pair, while \Abot encodes the answer it sampled,
  (4) \Qbot updates its state to $S^Q_t$, predicts an attribute pair via prediction LSTM at round $T$, and receives a reward.}
  \label{fig:models}
\end{figure*}
We formalize \Qbot and \Abot as agents operating in a partially observable world and optimize their policies using deep reinforcement learning.

\noindent\textbf{States and Actions.}
Each agent observes its own input (task $G$ for \Qbot and object instance $I$ for \Abot) and the output of the other agent as a stochastic environment.
At the beginning of round $t$, \Qbot observes state $s_Q^t{=}[G, q_1, a_1, \dots, q_{t-1}, a_{t-1}]$ and acts by uttering some token $q_t$ from its vocabulary $V_Q$.
Similarly, \Abot observes the history and this new utterance as state $s_A^t{=}[I, q_1, a_1, \dots, q_{t-1},a_{t-1},q_t]$ and emits a response $a_t$ from $V_A$.
At the last round, \Qbot takes a final action by predicting a pair of attribute values $\hat{w}^G = (\hat{w}^G_1, \hat{w}^G_2)$ to solve the task.

\noindent\textbf{Cooperative Reward.}
Both \Qbot and \Abot are rewarded identically based on the accuracy of \Qbot's prediction $\hat{w}^G$,
receiving a positive reward of $R{=}1$ if the prediction matches ground truth $w^G$ and a negative reward of $R{=}{-}10$ otherwise.
We arrive at these values empirically based on the speed of convergence in our experiments.

\noindent\textbf{Policy Networks.}
We model \Qbot and \Abot as operating under stochastic policies $\pi_Q(q_t{\mid}s^Q_t; \theta_Q)$
and $\pi_A(a_t{\mid}s^A_t; \theta_A)$ respectively, which we instantiate as
LSTM-based models.
\sk{We use lower case characters (\eg $s_t^Q$) to denote the strings (\eg \Qbot's token at round $t$), and upper case $S_t^Q$ to denote the corresponding vector as encoded by the model.}

As show in \reffig{fig:models}, \Qbot is modeled with three modules -- 
speaking, listening, and prediction. 
The task $G$ is received as a 6-dimensional one-hot encoding over the space 
of possible tasks and embedded via the listener LSTM.
\sml{At each round $t$, 
the \emph{speaker network} models the probability of output utterances $q_t\in V_Q$ based on  the state $S^Q_{t-1}$.
This is modeled as a fully-connected layer followed by a softmax that transforms $S^Q_{t-1}$ to a distribution over $V_Q$.}
After receiving the reply $a_t$ from \Abot, the 
\emph{listener LSTM} updates the state by processing both tokens of the  dialog exchange.
In the final round, the \emph{prediction LSTM} is unrolled twice to produce \Qbot's prediction based on the final state $S_T^Q$ and the task $G$.
\sk{As before, task $G$ is fed in as a one-hot encoding to the \sml{prediction LSTM} for two time steps, resulting in a pair of outputs used as the prediction} \sml{$\hat{w}_G$}.

Analogously, \Abot is modeled as a combination of 
a \emph{speaker network}, a \emph{listener LSTM}, and an \emph{instance encoder}.
Like in \Qbot, the speaker network \sml{models the probability of utterances $a_t\in V_A$ given the state $S_t^A$} and the listener LSTM updates the state $S_t^A$ based on dialog exchanges.
\sk{The instance encoder embeds each one-hot attribute vector via a linear layer and concatenates all three encodings to obtain a unified instance representation.}

\noindent\textbf{Learning Policies with REINFORCE.}
To train these agents, we update policy parameters $\theta_Q$ and $\theta_A$ using the popular REINFORCE \cite{williams1992simple} policy gradient algorithm.
Note that while the game is fully-cooperative, we do not assume full observability of one agent by another, opting instead to treat each agent as part of the stochastic environment when updating the other.
We will now derive the parameter gradients for our setup.

\sk{
Recall that our agents take actions -- utterances ($q_t$ and $a_t$) and attribute prediction ($\hat{w}_G$) -- and our
objective is to maximize the expected reward under the agents' policies:}

\vspace*{-20pt}
\begin{center}
\resizebox{1\columnwidth}{!}{
\begin{minipage}{\columnwidth}
{
\begin{subequations}
\begin{align}
\max_{\theta_A, \theta_Q}
J(\theta_A, \theta_Q) \quad \text{where,} \\
J(\theta_A, \theta_Q) =
\mathop{\mathbb{E}}_{\pi_Q, \pi_A} \left[ R \big(\hat w^G, w^G \big)  \right]
\end{align}
\end{subequations}}
\end{minipage}}
\end{center}

\vspace{-5pt}
\noindent 
\sk{
Though the agents receive the reward at the end of gameplay, all intermediate actions are assigned the same reward $R$.
Following the REINFORCE algorithm, we write the gradient of this expectation as an expectation of policy gradients.
For $\theta_Q$, we derive this explicitly at a time step $t$:}

\resizebox{0.85\columnwidth}{!}{
\centering
\begin{minipage}{1.2\columnwidth}
\begin{eqnarray}
\nabla_{\theta_Q}J
&{=}& \nabla_{\theta_Q} \left[\mathop{\mathbb{E}}_{\pi_Q, \pi_A} \left[ R \big(\hat w^G, w^G \big)  \right] \right]
\nonumber \\
&{=}& \nabla_{\theta_Q} \left[ \sum_{q_t,a_t}\pi_Q\left(q_t | s^Q_{t-1}\right)\pi_A\left(a_t | s^A_{t}\right) R(.)  \right] \nonumber \\
&{=}& \sum_{q_t,a_t}\pi_Q\left(q_t | s^Q_{t-1}\right)\nabla_{\theta_Q}\log\pi_Q\left(q_t | s^Q_{t-1}\right) \pi_A\left(a_t | s^A_{t}\right)R(.) \nonumber\\
&{=}& \mathop{\mathbb{E}}_{\pi_Q, \pi_A}\left[ R(.)  \, \nabla_{\theta_Q}\log\pi_Q\left(q_t | s^Q_{t-1}\right)  \right]
\hspace{-42pt}
\end{eqnarray}
\end{minipage}}\\[8pt]
Similarly, gradient \wrt $\theta_A$, \ie, $\nabla_{\theta_A}J$ will be:
{\small
\begin{align}
\vspace{-15pt}
\nabla_{\theta_A} J = \mathop{\mathbb{E}}_{\pi_Q, \pi_A}\left[ R(.) \, \nabla_{\theta_A}\log\pi_A\left(a_t | s^A_{t}\right) \right]
\vspace{-18pt}
\end{align}
}\vspace*{-15pt}

\sk{
As is standard practice, we estimate these expectations with sample averages -- sampling an environment (object instance and task), sampling a dialog between \Qbot and \Abot, culminating in a prediction from \Qbot and the received reward.
The REINFORCE update rule above has an
intuitive interpretation -- an \emph{informative} dialog $(q_t, a_t)$ that leads to positive reward will be made more probable (positive gradient), while a poor exchange leading to negative reward will be pushed down (negative gradient).}


\textbf{Implementation Details.}
All our models are implemented using the Pytorch\footnote{\href{http://github.com/pytorch/pytorch}{\texttt{github.com/pytorch/pytorch}}} deep learning framework.
\sk{To represent instances, we learn a $20$ dimensional embedding for every possible attribute values and concatenate the three instance attributes to obtain a final instance representation of size $60$.}
Tokens from $V_Q$ and $V_A$ are encoded as one-hot vectors and then embedded into $20$ dimension vectors.
Both \Abot and \Qbot learn their own token embeddings without sharing.
The listener networks in both agents are implemented as LSTMs with a hidden layer size of $50$ dimensions.
All modules within an agent are initialized using the Xavier method \cite{glorot2010understanding}.

We use $1000$ episodes of two-round dialogs to compute policy gradients, and perform updates according to Adam optimizer \cite{kingma_iclr15}, with a learning rate of $0.01$.
Furthermore, gradients are clipped at $[-5.0, 5.0]$.
For faster convergence, $80\%$ of train episodes for the next iteration are from instances misclassified by the current network, while randomly sampling the remaining from all instances.
Our code is publicly available\footnote{\href{http://github.com/batra-mlp-lab/lang-emerge}{\texttt{github.com/batra-mlp-lab/lang-emerge}}}.

\vspace{\sectionReduceTop}
\section{The Road to Compositionality}
\label{sec:game-config}
\vspace{\sectionReduceBot}

    This section details our key observation -- 
	that while the agents always successfully invent 
    a language to solve the game with
near-perfect accuracies, the invented languages are decidedly \emph{not} compositional, interpretable, or `natural'
(\eg \Abot ignoring \Qbot's utterances and simply 
encoding every object with a unique symbol 
if the vocabulary is sufficiently large).
In our setting, the language being compositional simply amounts to the ability of the agents to communicate the compositional atoms of a task (\eg \textit{shape} or \textit{color}) and an instance (\eg \textit{square} or \textit{blue}) independently.

    
    Through this section, we present a series of settings that get progressively more restrictive to coax the agents towards adopting a compositional language, providing analysis of the learned languages \sk{and `cheating' strategies } developed along the way.
    \reftab{tab:overview-setting} summarizes results for all settings.
    In all experiments, optimal policies (achieving near-perfect \sml{training} rewards) were found. 
    For each setting, we provide qualitative analysis of the learned languages and report their ability to generalize to unseen instances. 
    We use 80\% of the object-instances for training and the remaining 20\% for evaluation.
    


\vspace{\subsectionReduceTop}
\subsection{Overcomplete Vocabularies}
\label{sec:overcomplete}
\vspace{\subsectionReduceBot}

\begin{figure}
	\includegraphics[width=\columnwidth]{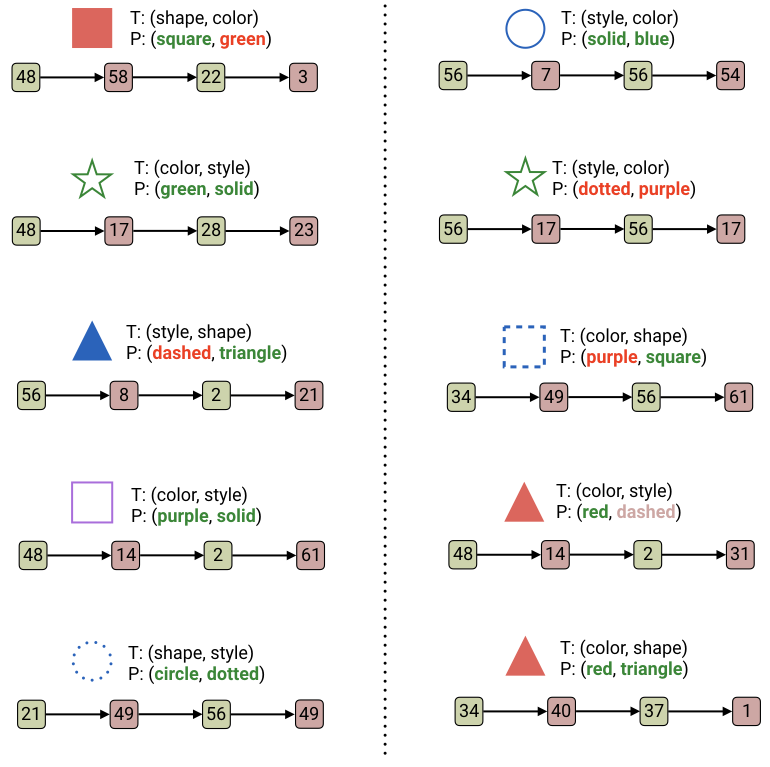}\\
    \caption{
    \sk{Overcomplete vocabularies setting ($|V_Q| = |V_A| =64$, \refsec{sec:overcomplete}).
    Owing to a large vocabulary, we denote the tokens using numbers, as opposed to English alphabet characters shown in other figures.
    \Abot mostly ignores what \Qbot asks and instead conveys the instance using pairs of symbols across rounds unique to an instance, leading to a highly non-human intuitive and non-compositional language.}}
    \label{fig:ex_overcomplete}
\end{figure}

\sk{
We begin with the simplest setting where both \Abot and \Qbot are given arbitrarily large 
vocabularies.
We find that when $|V_A|$ is greater than the number of instances (64), the learned 
policy mostly ignores what \Qbot asks and instead has \Abot convey the instance using pairs of symbols across rounds unique to an instance, 
\eg, both token pairs $(a_1, a_2) {=} (14, 31), (40, 1)$ convey \emph{(red, triangle, filled)}, as shown in \reffig{fig:ex_overcomplete}.
Notice, this means no `dialog' is necessary and amounts to each agent having a codebook that maps symbols to object instances.
In essence, this setting has collapsed to an analog of Lewis Signaling (LS)} \sml{ game with \Abot signaling its complete world state and \Qbot simply reporting the target attributes.}
More examples to illustrate this behavior for this setting are shown in \reffig{fig:ex_overcomplete}.

Perhaps as expected, the generalization of this language to unseen instances is quite poor (success rate: 25.6\%).
The adopted strategy of mapping instances to token pairs fails for test instances containing novel combinations of attributes for which the agents lack an agreed-upon code from training.

It seems clear that like in human communication \cite{novak_nature00}, a limited vocabulary that cannot possibly encode the richness of the world seems to be necessary for non-trivial dialog to emerge. We explore such a setting next.



\vspace{\subsectionReduceTop}
\subsection{Attribute-Value Vocabulary}
\label{sec:Mem-L}
\vspace{\subsectionReduceBot}

\begin{figure}
    \includegraphics[width=\columnwidth]{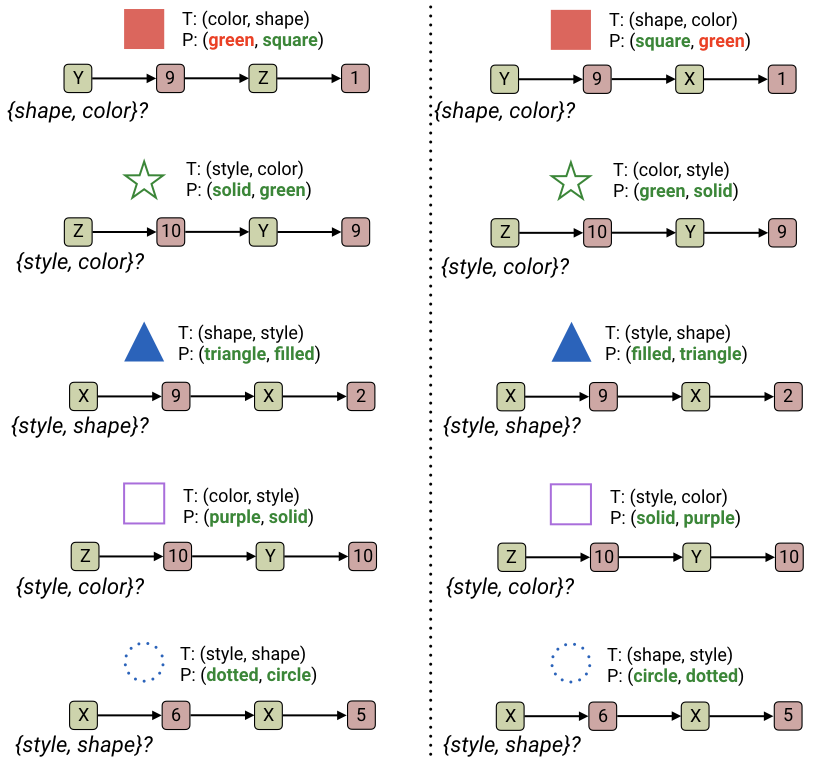}\\
    \caption{
    \sk{Attribute and Value vocabulary setting ($|V_Q| = 3, |V_A| = 12$, \refsec{sec:Mem-L}).
    We show symmetric tasks for each instance on either side to illustrate the similarities in the language between the agents.
    As seen here, \Qbot maps symmetric tasks in an order-agnostic fashion, and uses only the first token to convey task information to \Abot.}}
    \label{fig:ex_attribute_vocab}
\end{figure}

Since our world has 3 attributes (\emph{shape}/\emph{color}/ \emph{style}) and $4+4+4=12$ possible settings of their states, 
one may believe that the intuitive 
choice of $|V_Q|=3$ and $|V_A|=12$ will be enough to circumvent 
the `cheating' enumeration strategy from 
the previous experiment. 
Surprisingly, we find that the new 
language learned in this setting is not only 
decidedly non-compositional but also very difficult to interpret! 
Some observations follow from \reffig{fig:ex_attribute_vocab} that shows sample dialogs for this setting.

\sk{
We observe that \Qbot uses only the first round to convey the task to \Abot by encoding tasks in an order-agnostic fashion, as:
\emph{(style, shape),(shape, style)} ${\rightarrow}$ X,
\emph{(color, shape),(shape, color)} ${\rightarrow}$ Y, and
\emph{(color, style),(style, color)} ${\rightarrow}$ Z.}
\sk{Thus, multiple rounds of utterance from \Qbot are rendered unnecessary and we find the second round is inconsistent across instances even for the same task.
For example, symmetric tasks \emph{(color, shape)} and \emph{(shape, color)} from first row of \reffig{fig:ex_attribute_vocab} induce $q_1{=}Y$ as the first token from \Qbot.}

Given the task from \Qbot in the first round, \Abot only needs to identify one of the $4{\times}4{=}16$ attribute pairs for a given task.
Rather than ground its symbols into individual states, \Abot follows a `set partitioning' strategy, \ie 
\Abot identifies a pair of attributes with a unique combinations of round 1 and 2 utterances (\ie the round 2 utterance has no meaning independent from round 1). 
Thus, symbols are reused across tasks to describe different  attributes (\ie symbols do not have individual consistent groundings). This `set partitioning' strategy is consistent 
with known results from game theory on Nash equilibria in 
`cheap talk' games~\cite{crawford_econ82}. 

This strategy has improved generalization to unseen instances  because it is able to communicate the task; however, it fails on unseen attribute value combinations because 
it is not compositional.

\vspace{\subsectionReduceTop}
\subsection{Memoryless \Abot, Minimal Vocabulary}
\label{sec:nomem_min}
\vspace{\subsectionReduceBot}
\begin{figure}
	\centering
	\includegraphics[width=\columnwidth]{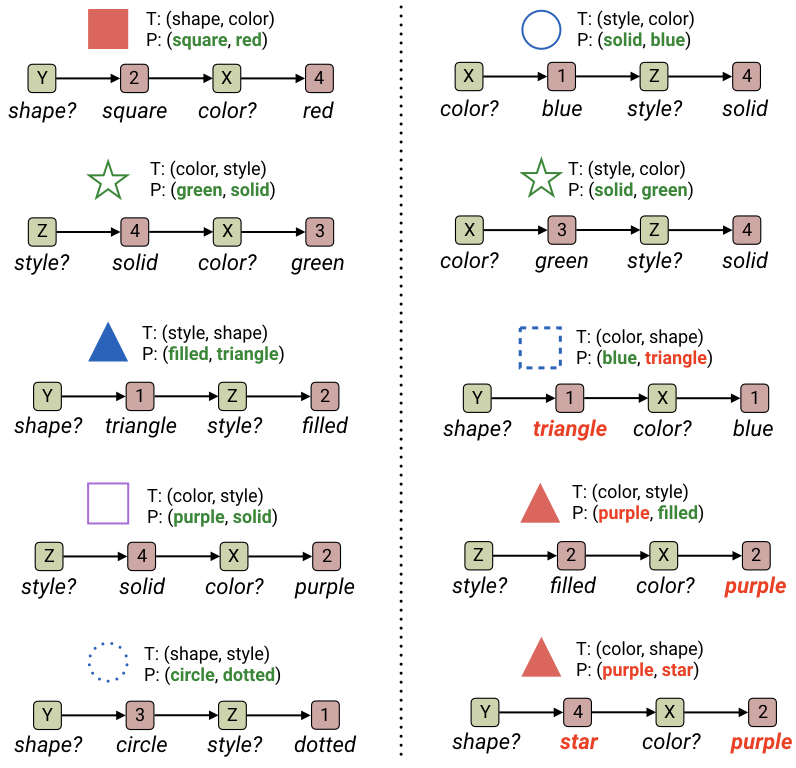}
    \caption{
    \sk{Example dialogs for memoryless \Abot, minimal vocabulary setting ($|V_Q| = 3, |V_A| = 4$, \refsec{sec:nomem_min}).
    Learnt language is consistent and grounded, denoted below each token.
    Incorrect predictions on unseen instances (right, bottom) are also shown.
    Notice that the incorrectly predicted attribute is still from the right category (a color attribute for \emph{color}).}}
    \label{fig:ex_minimal_nomem}
\end{figure}
The key problem with the previous setting is that \Abot's utterances \emph{mean different things} based on the round of dialog ($a_1=1$ is different from $a_2=1$).
Essentially, the communication protocol is over-parameterized and we must limit it further. 
First, we limit \Abot's vocabulary to $|V_A|{=}4$ to reduce the number of `synonyms' the agents learn.
Second, 
we eliminate \Abot's capability to identify different rounds of interaction by \emph{removing \Abot's memory}.
In other words, we reset the state vector $S^A$ at each time step so that \Abot can no longer distinguish rounds from one another.
\sk{
By doing so, we hypothesize that \Qbot must now ground its own and \Abot's tokens consistently across rounds to be able to communicate with a memoryless \Abot.}

These restrictions result in a learned language 
that grounds \emph{individual symbols} into 
attributes and their states. 
For example, \Qbot learns that $Y\rightarrow\mbox{shape}$, $X\rightarrow\mbox{color}$, and $Z\rightarrow\mbox{style}$. 
\Qbot does not however learn to always utter these symbols in the same order as the task, \eg asking for \emph{shape} first for both $(color, shape)$ and $(shape, color)$.
Notice that this is perfectly valid as \Qbot can later re-arrange the attributes in the task desired order.
Similarly, \Abot learns mappings to attribute values for each attribute query that remain consistent regardless of round (\ie when asked for $color$, $1$ always means $blue$).

This is similar to learned languages reported in recent works and is most closely related to \citet{visdial_rl}, who solve this problem by taking away \Qbot's state rather than \Abot's memory.
Their approach of taking away task $G$ from \Qbot's state can be interpreted as \Qbot `forgetting' the task after interacting with \Abot.
However, this behavior of \Qbot to remember the task only during dialog but not while predicting is somewhat unnatural compared to our setting.

\reftab{tab:both_grounding_minimal} enumerates the learnt groundings for both the agents.
Given this mapping, we can predict a plausible dialog between the agents for any unseen instance and task combination.
Notice that this is possible only due to the compositionality in the emergent language between the two agents.
For example, consider solving the task \emph{(shape, color)} for an instance \emph{(red, square, filled)} from \reffig{fig:lang_chart}(b).
\Qbot queries \emph{Y (shape)} and \emph{X (color)} across two rounds, and receives \emph{2 (square)} and \emph{4 (red)} as answers.
More examples along with grounded meaning of each tokens are shown in \reffig{fig:ex_minimal_nomem}.

\begin{table}
\setlength{\tabcolsep}{3pt}
 \resizebox{0.55\linewidth}{!}{
\begin{subtable}{0.6\linewidth}
	\centering
    \begin{tabular}{c c c c }
 	\toprule
   & \multicolumn{3}{c}{\textbf{Attributes}} \\ \cline{2-4}
   & \emph{color} & \emph{shape} & \emph{style}\\
   $V_A$ & \emph{X} & \emph{Y} & \emph{Z}\\
  \midrule
	\emph{1} & \emph{blue} & \emph{triangle} & \emph{dotted}\\
    \emph{2} & \emph{purple} & \emph{square} & \emph{filled}\\
    \emph{3} & \emph{green} & \emph{circle} & \emph{dashed}\\
    \emph{4} & \emph{red} & \emph{start} & \emph{solid} \\
    \bottomrule
  \end{tabular}
  \caption{\Abot}
  \label{tab:abot_grounding_minimal}
\end{subtable}}
\resizebox{0.361\linewidth}{!}{
\begin{subtable}{0.39\linewidth}
	\centering
	\begin{tabular}{c c}
  \toprule
  \textbf{Task} & \textbf{$q_1, q_2$}\\
  \midrule
    \emph{(color, shape)} & \multirow{2}{*}{\emph{Y, X}}\\
    \emph{(shape, color)} &\\
	\emph{(shape, style)} & \multirow{2}{*}{\emph{Y, Z}}\\
    \emph{(style, shape)} &\\
    \emph{(color, style)} & \emph{Z, X}\\
    \emph{(style, color)} & \emph{X, Z}\\
  \bottomrule
  \end{tabular}
  \caption{\Qbot}
  \label{tab:qbot_grounding_minimal}
\end{subtable}}
\caption{
\sk{Emergence of compositional grounding for language learnt by the agents.
\textbf{\Abot} (\reftab{tab:abot_grounding_minimal}) learns consistent mapping across rounds, depending on the query attribute.
Token grounding for \textbf{\Qbot} (\reftab{tab:qbot_grounding_minimal}) depends on the task at hand.
Though compositional, \Qbot does not necessarily query attribute in the order of task, but instead re-arranges accordingly at prediction time as it contains memory.}}
\label{tab:both_grounding_minimal}
\end{table}

Intuitively, this consistently grounded and compositional language has the greatest ability to generalize among the settings we have explored, correctly answering 74.4\% of the held out instances.
We note that errors in this setting seem to largely be due to \Abot giving an incorrect answers despite \Qbot asking the correct questions to accomplish the task.
A plausible reason could be the model approximation error stemming from the instance encoder as test instances are unseen and have novel attribute combinations.

\begin{table*}
	\centering
	\setlength{\tabcolsep}{4pt}
    \resizebox{0.98\textwidth}{!}{
	\centering
    \begin{tabu}{p{0.82in} c c c c c c c c c}
    	\toprule
        \multirow{2}{*}{Setting} 
    	& \multicolumn{2}{c}{Vocab.} 
        & \multicolumn{2}{c}{Memory} 
        & \multicolumn{2}{c}{Seen (\%)}
        & \multicolumn{2}{c}{Unseen (\%)}
        & \multirow{2}{*}{Characteristics}\\
\cmidrule(lr){2-3} \cmidrule(lr){4-5} \cmidrule(lr){6-7} \cmidrule(lr){8-9}
  	  & \textbf{$V_Q$} & \textbf{$V_A$} & A & Q & Both & One & Both & One \\ \midrule

     Overcomplete (\S\ref{sec:overcomplete}) & 64 & 64 & \cmark & \cmark & 100 & 100 & 25.6 & 79.5
    	&  
  		\begin{minipage}[c]{0.5\textwidth}
          \begin{compactitem}
          \item Non-compositional language
          \item \Qbot insignificant
          \item Inconsistent \Abot grounding across rounds
          \item Poor generalization to unseen instances
          \end{compactitem}
        \end{minipage}\\ \midrule         
        Attr-Value (\S\ref{sec:Mem-L}) & 3 & 12 & \cmark & \cmark & 100 & 100 & 38.5 & 88.4
        &
   		\begin{minipage}[c]{0.5\textwidth}
          \begin{compactitem}
          \item Non-compositional language
          \item \Qbot uses one round to convey task
          \item Inconsistent \Abot grounding across rounds
          \item Poor generalization to unseen instances
          \end{compactitem}
        \end{minipage}\\ \midrule
        
        NoMem-Min (\S\ref{sec:nomem_min}) & 3 & 4 & \xmark & \cmark & 100 & 100 & \textbf{74.4} & \textbf{94.9}
        &
        \begin{minipage}[c]{0.5\textwidth}
          \begin{compactitem}
          \item Compositional language
          \item \Qbot uses both rounds to convey task
          \item Consistent \Abot grounding across rounds
          \item Good generalization to unseen instances
          \end{compactitem}
        \end{minipage}\\ \bottomrule
    \end{tabu}}
    \caption{Overview of settings we explore to analyze the language learnt by two agents in a cooperative game, \tnt. Last two columns measure generalization in terms of prediction accuracy of \textbf{both} or at least \textbf{one} of the attribute pair, on a held-out test set containing unseen instances.}
    \label{tab:overview-setting}
\end{table*}
\vspace{\sectionReduceTop}
\section{Evolution of Language}
\vspace{\sectionReduceBot}
As demonstrated by the previous sections, even though compositional language is one of the optimal policies, the agents tend to learn other equally useful forms of communication.
Thus, compositional language does not naturally emerge without an explicit need for it.
Even in situations where compositionality does emerge, perhaps it is more interesting to analyze the process of emergence than the learnt language itself.
Therefore, we present such a study that explicitly identifies \emph{when} each symbol has been grounded by the agents in the training timeline, along with implications thereof on the performance on \tnt game.
\subsection{Dialog Trees}
\label{sec:dialog_trees_supp}

\begin{figure}
	\centering
	\includegraphics[width=0.99\columnwidth]{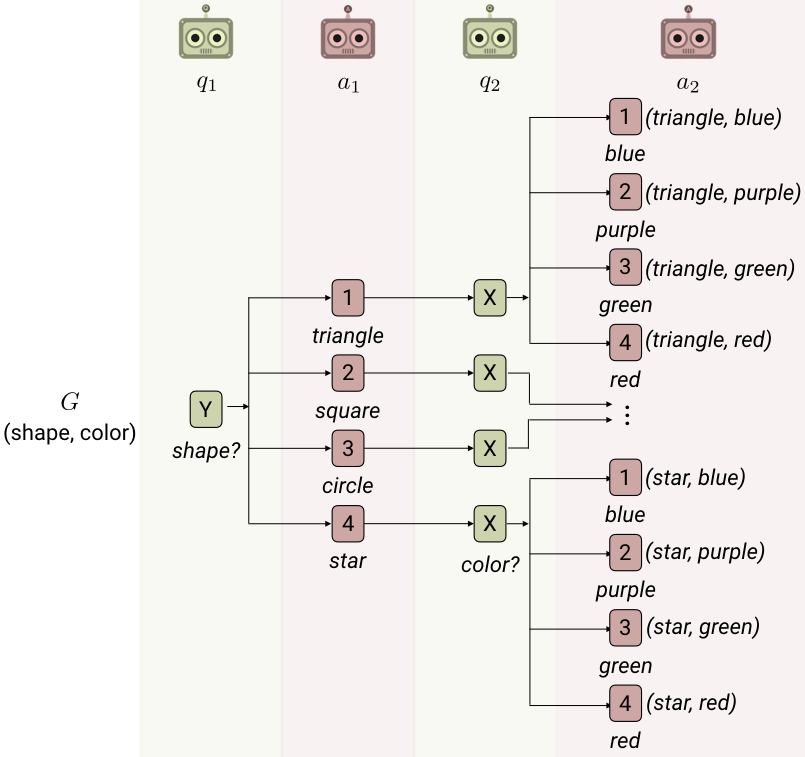}
    \caption{
    \sk{Dialog tree for memoryless \Abot and minimal vocabulary setting (Sec.4.3), shown only for one task \emph{(shape, color)}.
    Every dialog between the agents results in a tree traversal beginning from the root, \eg, $Y{\rightarrow}1{\rightarrow}X{\rightarrow}1$ lands us in the top-right leaf. See text for more details.}}
    \label{fig:dialog_tree_example}
\end{figure}

When two agents--\Qbot and \Abot--converse with each other, they can be seen as traversing through a dialog tree, a subtree of which is depicted in \reffig{fig:dialog_tree_example}.
Simply put, a dialog tree is an enumeration of all possible dialogs represented in the form of tree, with levels of the tree corresponding to the round of interaction.
To elaborate, consider a partial dialog tree for \emph{(shape, color)}  task shown in \reffig{fig:dialog_tree_example} for the setting in \refsec{sec:nomem_min}.
For \Qbot's first token $q_1=Y$, \Abot has $|V_A|=4$ plausible replies shown as a 4-way branch off.
In general, the dialog tree for \tnt contains a total of $|V_Q|^2|V_A|^2$ leaves and is 4 levels deep.
We use the dialog between the agents to descend and land in one of these leaves.

Dialog trees offer an interesting alternate view of our learning problem.
The goal of learning communication between the two agents can be equivalently seen as mapping \emph{(instance, task)} pairs to one of the dialog tree leaves.
Each leaf is labeled with an attribute pair used to accomplish the prediction task.
For example, if solving \emph{(shape, color)} for \emph{(blue, triangle, solid)} results in the dialog $Y {\rightarrow} 1 {\rightarrow} X {\rightarrow} 1$,
we descend the dialog tree along the corresponding path and assign the tuple \emph{(blue, triangle, solid, shape, color)} to the resulting leaf.
In case of a compositional, grounded dialog, all tuples of the form \emph{(blue, triangle, $\ast$, shape, color)} would get mapped to the same leaf, which can then be labeled as \emph{(triangle, blue)} to successfully solve the task.
Note the wildcard \emph{style} attribute in the tuple above, as it is irrelevant for this particular task.

In the following section, we use dialog trees to explore the evolution of language as learnt by the two agents in the memoryless A-bot, minimal vocabulary setting in \refsec{sec:nomem_min}.

\begin{figure*}
	\centering
	\includegraphics[width=0.99\textwidth]{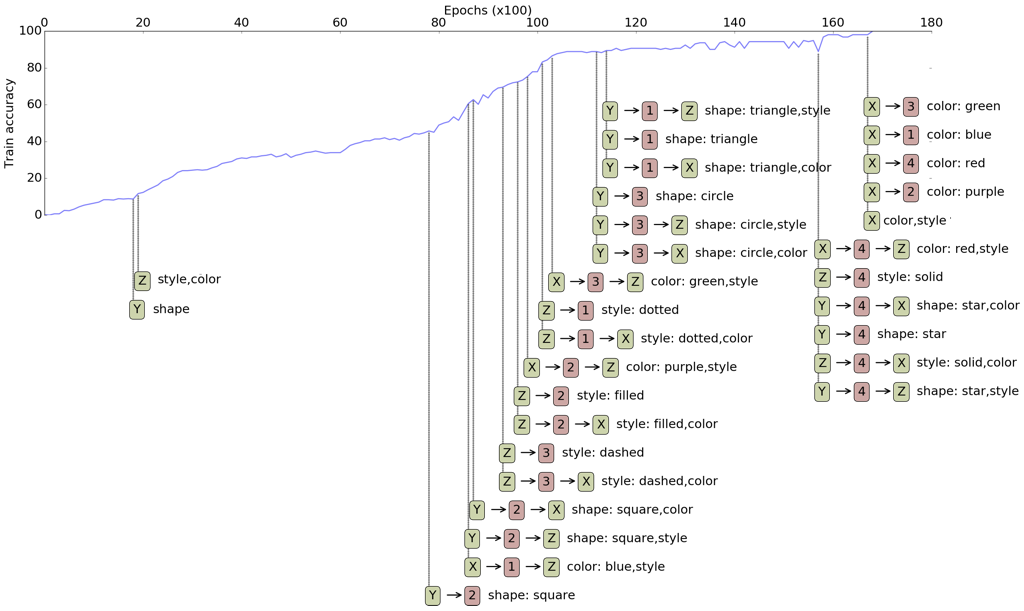}
    \caption{Evolution of Language: timeline shows groundings learned by the agents during training, overlaid on the accuracy. Note that \Qbot learns encodings for all tasks early (around epoch $20$) except \emph{(style, color)}. Improvement in accuracy is strongly correlated with groundings learnt.}
    \label{fig:lang_chart}
    \vspace*{-4pt}
\end{figure*}

\subsection{Evolution Timeline}
\label{sec:evol_timeline}
To gain further insight into the languages learned, 
we create a \emph{language evolution} plot shown in \reffig{fig:lang_chart}. 
Specifically, at 
regular intervals during policy learning, we construct dialog trees. 
At some point in the learning, the nodes in the tree become and stay `pure' 
(all \emph{(instance, task)} at the node are identical), 
at which point we can say that the agents have learned this dialog subsequence. 
\reffig{fig:lang_chart} depicts a timeline of concepts learned at various nodes of the trees during training.
We next describe the procedure to identify when a particular `concept' has been grounded by the agents in their language.

\sk{
\textbf{Construction.} 
After constructing dialog trees at regular intervals, we identify `concepts' at each node/leaf using the dialog tree of the completely trained model, which achieves a perfect accuracy on train set.
A concept is simply the common trend among all the \emph{(instance, task)} tuples either assigned to a leaf or contained within the subtree with a node as root.
To illustrate, the concept of the top right leaf in \reffig{fig:dialog_tree_example} is \emph{(blue, triangle, $\ast$, shape, color)}, \ie, all instances assigned to that leaf for \emph{(shape, color)} task are blue triangles.
Next, given a resultant concept for each of the node/leaf, we backtrack in time and check for the first occurrence when only tuples which satisfy the corresponding concept are assigned to that particular node/leaf.
In other words, we compute the earliest time when a node/leaf is `pure' with respect to its final learned concept.
Finally, we plot these leaves/nodes and the associated concept with their backtracked time to get \reffig{fig:lang_chart}.}

\textbf{Observations.} We highlight the key observations from \reffig{fig:lang_chart} below:
\begin{compactenum}[(a)]\labelwidth=2em

\item The agents ground most of the tasks initially at around epoch 20.
Specifically, \Qbot assigns \emph{Y} to both \emph{(shape, style), (style, shape), (shape,color)} and \emph{(color, shape)}, while \emph{(color, style)} is mapped to \emph{Z}.
Hence, \Qbot learns its first token very early into the training procedure at around 20 epochs.

\item The only other task \emph{(style, color)} is grounded towards the end (around epoch 170) using \emph{X}, leading to an immediate convergence.

\item We see a strong correlation between improvement in performance and when agents learn a language grounding.
In particular, there is an improvement from $40\%$ to $80\%$ within a span of 25 epochs where most of the grounding is achieved, as seen from \reffig{fig:lang_chart}.
\end{compactenum}

\vspace{-8pt}
\vspace{\sectionReduceTop}
\section{Conclusion}
\vspace{\sectionReduceBot}
\vspace*{-10pt}
In conclusion, 
we presented a sequence of `negative' results 
culminating in a `positive' one -- showing
that while most invented languages are effective 
(\ie achieve near-perfect rewards), they 
are decidedly \emph{not} interpretable or compositional.
Our goal is simply to improve understanding and interpretability 
of invented languages in multi-agent dialog, place recent work in context, and inform fruitful directions for future work. 


{
\bibliographystyle{emnlp_natbib}
\bibliography{strings,main}
}
\end{document}